\documentclass[conference]{IEEEtran}
\IEEEoverridecommandlockouts
\usepackage{cite}
\usepackage{amsmath,amssymb,amsfonts}
\usepackage{algorithmic}
\usepackage{graphicx}
\usepackage{textcomp}
\usepackage{xcolor}
\def\BibTeX{{\rm B\kern-.05em{\sc i\kern-.025em b}\kern-.08em
    T\kern-.1667em\lower.7ex\hbox{E}\kern-.125emX}}
\begin{document}

\title{DADgraph: A Discourse-aware Dialogue Graph Neural Network for Multiparty Dialogue Machine Reading Comprehension}

\author{\IEEEauthorblockN{Jiaqi Li}
\IEEEauthorblockA{
\textit{Harbin Institute of Technology}\\
Harbin, China \\
jqli@ir.hit.edu.cn}
\and
\IEEEauthorblockN{Ming Liu}
\IEEEauthorblockA{
\textit{Harbin Institute of Technology}\\
Harbin, China \\
mliu@ir.hit.edu.cn}
\and
\IEEEauthorblockN{Zihao Zheng}
\IEEEauthorblockA{
\textit{Harbin Institute of Technology}\\
Harbin, China \\
zhzheng@ir.hit.edu.cn}
\and
\IEEEauthorblockN{Heng Zhang}
\IEEEauthorblockA{
\textit{Harbin Institute of Technology}\\
Harbin, China \\
hzhang@ir.hit.edu.cn}
\and
\IEEEauthorblockN{Bing Qin* \thanks{\ \*Corresponding author.}
}
\IEEEauthorblockA{
\textit{Harbin Institute of Technology}\\
Harbin, China \\
qinb@ir.hit.edu.cn}
\and
\IEEEauthorblockN{Min-Yen Kan}
\IEEEauthorblockA{
\textit{National University of Singapore}\\
Singapore, Singapore \\
kanmy@comp.nus.edu.sg}
\and
\IEEEauthorblockN{Ting Liu}
\IEEEauthorblockA{
\textit{Harbin Institute of Technology}\\
Harbin, China \\
tliu@ir.hit.edu.cn}
}

\maketitle

\begin{abstract}

Multiparty Dialogue Machine Reading Comprehension (MRC) differs from traditional MRC as models must handle the complex dialogue discourse structure, previously unconsidered in traditional MRC.
To fully exploit such discourse structure in multiparty dialogue, we present a discourse-aware dialogue graph neural network, {\it DADgraph}, which explicitly constructs the dialogue graph using discourse dependency links and discourse relations. To validate our model, we perform experiments on the {\it Molweni} corpus, a large-scale MRC dataset built over multiparty dialogue annotated with discourse structure. 
Experiments on Molweni show that our discourse-aware model achieves statistically significant improvements compared against strong neural network MRC baselines. 
\end{abstract}


\section{Introduction}

Research into multiparty dialogue has grown rapidly given the growing ubiquity of dialogue agents~\cite{shi2019deep,GSNijcai2019,li2019keep,zhao2019abstractive,sun2019dream,N16IntegerLP,D15DP4MultiPparty}. The machine-aided comprehension of such dialogue, in the form of multiparty dialogue machine reading comprehension (MRC), has subsequently begun to attract research \cite{ma2018challenging,sun2019dream,yang2019friendsqa}. 


Work on general machine reading comprehension is flourishing. Most existing datasets for general machine reading comprehension adopt well-written prose passages and historical questions as inputs \cite{richardson2013mctest,rajpurkar2016squad,lai2017race,choi2018quac,reddy2019coqa}. In inputs for such general MRC, a passage is a continuous text where there is a discourse relation between every pair of adjacent sentences. Therefore, we can regard each paragraph in a passage as a linearly structured discourse.  In contrast, MRC for multiparty dialogue must consider the more complex, graphical nature of discourse structure: coherence between adjacent utterances is not a given; there may be no discourse relation between adjacent utterances. The discourse structure in such multiparty dialogues can be regarded as a dependency graph, where nodes are utterances.

\begin{figure*}
    \centering
    \includegraphics[width=\textwidth]{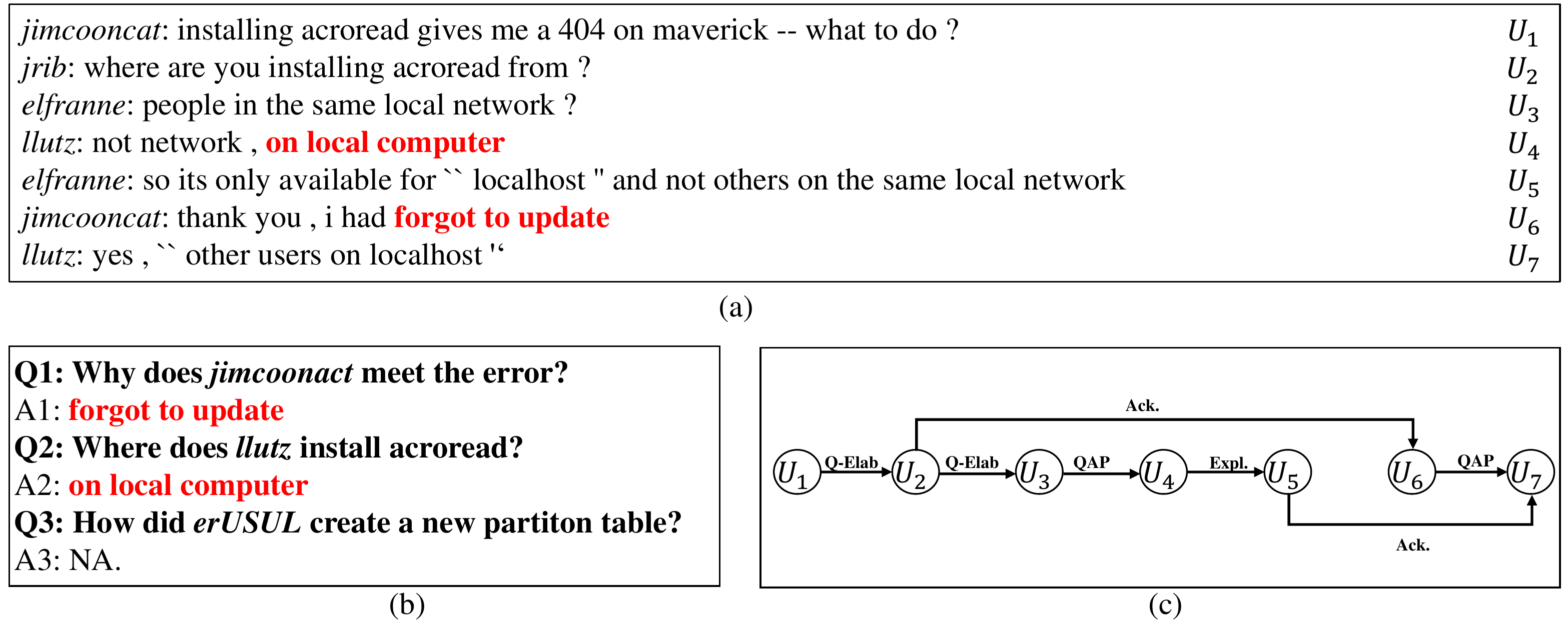}
    \caption{ (a) Multiparty dialogue from {\it Molweni}, with accompanying (b) contributed questions and answers, and (c) discourse structure. Correct answers are marked in red. Q-Elab, QAP, Expl and Ack. respectively represent the Question-Elaboration, Question-Answer Pair, Explanation and Acknowledgements relations.}
    \label{fig:example}
\end{figure*}

Figure~1 shows a multiparty dialogue example and its discourse structure from the {\it Molweni} dataset (\S~\ref{sec:exp}), where four speakers converse over seven utterances. The annotators of Molweni have contributed three questions (Fig.~\ref{fig:example}, b): two answerable ones (Q1 and Q2) and one unanswerable one (Q3).  They also have hand-annotated the discourse structure (Fig.~\ref{fig:example}, c), where nodes and edges represent utterances and their associated discourse relations, respectively.  We observe that adjacent utterance pairs can be incoherent, illustrating the key challenge.  
It is non-trivial to detect discourse relations, especially between non-adjacent utterances; and crucially, difficult to correctly interpret a multiparty dialogue without a proper understanding of the input's complex structure.  \\

\noindent \textbf{Hypothesis:} {\it Discourse structure informs multiparty dialogue MRC performance in modeling long-term dependencies.} \\

Discourse structure has been successfully applied to question answering and machine reading comprehension \cite{chai2004discourse,sun2007discourse,jansen2014discourse,narasimhan2015machine,sachan2015learning}. To the best of our knowledge, there is no prior work introducing discourse structure to multiparty dialogue MRC; i.e., all works on dialogue MRC do not consider the characteristic properties of multiparty dialogue. 


To utilize the discourse structure of multiparty dialogues, we propose {\it DADgraph}, a Discourse-Aware Dialogue graph convolutional network consisting of three key components. The first component is sequential context encoding which aims to learn the sequence structure of utterances. The second component is dialogue graph modeling. To effectively model multiparty dialogue discourse structure, we adopt graph neural networks. The third component is the MRC module. After processing the input through two different dialogue encoders, we feed the resultant dialogue representations to the MRC module to find the answer span. In contrast to the basic DialogueGCN~\cite{ghosal2019dialoguegcn} which uses a windowed context, our model represents the dialogue graph using discourse dependency links and discourse relations.




To the best of our knowledge, the are two dialogue MRC datasets, including the FriendsQA \cite{yang2019friendsqa} dataset and the Molweni dataset\cite{li2020molweni}. FriendsQA deriving from the \textit{Friends} TV show, comprises of 1,222 dialogues and 10,610 questions. 
However, the FriendsQA dataset lacks discourse structure annotation and does not directly serve to validate our hypotheses. 
As such, Molweni is more suited as it incorporates multiparty dialogue MRC corpus with discourse structure. For this reason, we only adopt the Molweni multiparty dialogue dataset, a large-scale span-based machine reading comprehension dataset. Molweni contains 10,000 dialogues with 88,303 utterances and 30,066 questions, inclusive of both answerable and unanswerable questions. Crucially, the Molweni dataset annotated its discourse relations -- all 78,245 present -- in all of its dialogues.


On Molweni, our discourse-aware graph model achieves state-of-the-art results compared with traditional MRC models including BiDAF \cite{seo2016BiDAF}, DocQA \cite{clark2018simple}, and BERT\cite{devlin2019bert}. DADgraph also outperforms the DialogueRNN \cite{majumder2019dialoguernn} and DialogueGCN \cite{ghosal2019dialoguegcn} dialogue-based models.  
\section{Related work}

Our work intersects MRC, discourse parsing and dialogue systems.  We review these areas with a focus on the choice of MRC dataset, as it is a critical aspect that enables the modeling in DADgraph.

\paragraph{Machine reading comprehension} MRC asks a system to answer questions with respect to an input passage. There are several types of datasets for machine comprehension, such as multiple-choice datasets \cite{richardson2013mctest,lai2017race}, answer sentence selection datasets \cite{wang2007jeopardy,yang2015wikiqa} and extractive datasets \cite{rajpurkar2016squad,joshi2017triviaqa,trischler2017newsqa,rajpurkar2018know}. 

\paragraph{Discourse parsing for multiparty dialogues} Discourse parsing for multiparty dialogues is a challenging task which aims to obtains the discourse dependency links and discourse relations between utterances. 
STAC \cite{asher2016discourse} and Molweni \cite{li2020molweni} are existing corpora for the task. 
The senses of discourse relation are introduced in \S~\ref{sec:method}. 
Most existing methods using traditional statistical machine learning models \cite{D15DP4MultiPparty,N16IntegerLP}, and more neural-based models for the task are still should be explored \cite{shi2019deep}.


\paragraph{Dialogue systems} Dialogue systems have achieved a great process with introducing deep learning.  \cite{wu2020diverse} \cite{young2020dialogue} and \cite{liu2019zero} respectively introduce commonsense knowledge, audio context and transferable latent variables into dialogue systems. \cite{ma2020survey} summarizes the literature on empathetic dialogue systems. The usage of discourse structure and topic information for dialogue generation and dialogue summarization would be a meaningful research problem. 

\section{Task definition}



\begin{figure*}
    \centering
    \includegraphics[width=0.95\textwidth]{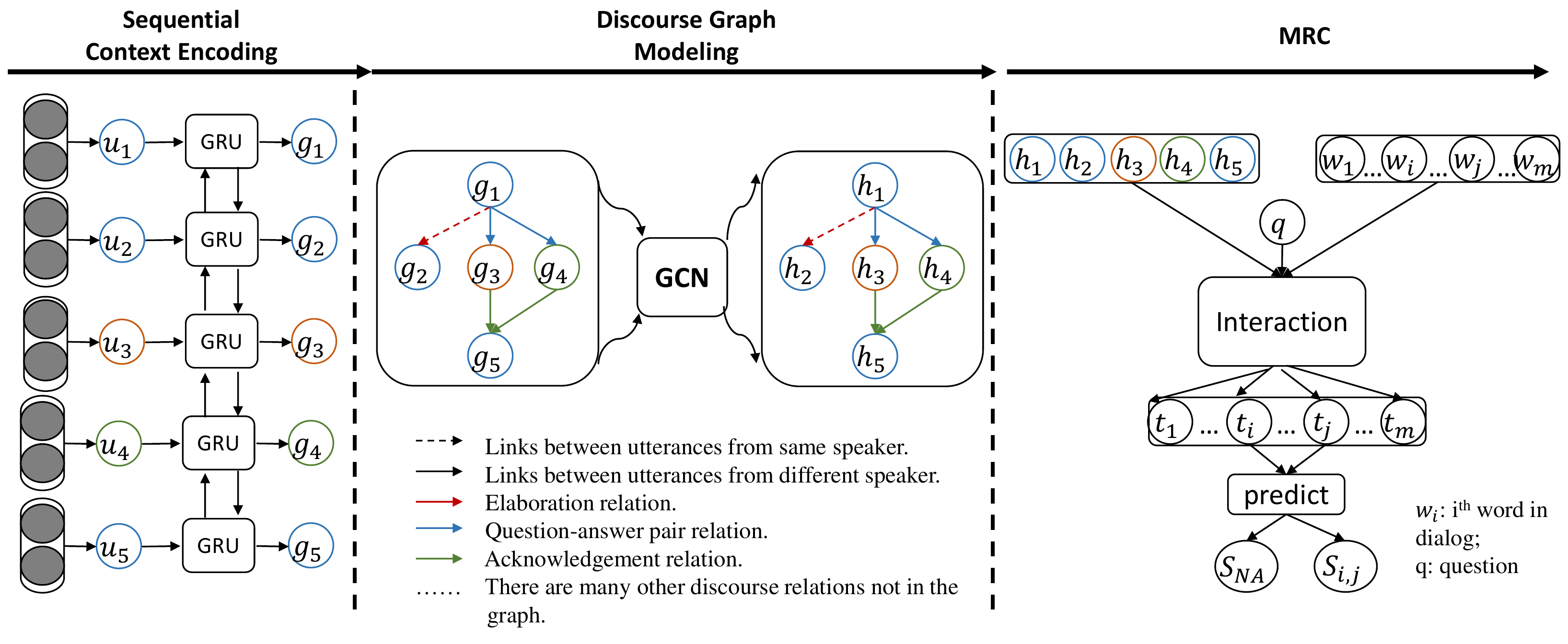}
    \caption{ Dialogue graph modeling using discourse structure. The edge between vertices is the discourse dependency link with discourse relation. Different colors of nodes and edges respectively represent different speakers and different discourse relations.}
    \label{fig:architecture}
\end{figure*}

Given a multiparty dialogue $d=\{u_1,u_2,...,u_N\}$ with $N$ utterances and $M$ questions $q=\{q_1,q_2,...,q_M\}$, the task is to predict answers $a=\{a_1,a_2,...,a_M\}$ for each question. Each utterance $u_i=\{s_i,c_i\}$ contains two parts: speaker $s_i$ and content $c_i$. Besides, all utterances are concatenated to get $d_{cat}$. There are two types of questions: answerable questions and unanswerable questions. If the question $q_i$ is answerable, the answer $a_i$ should be a continuous span in $d_{cat}$ including the index of start $S$ and end $E$ of the answer. Otherwise, the answer $a_i$ should be $NA$ (unanswerable).
$$a_i=\left\{
\begin{array}{lr}  
(S, E), & if \ q_i \ is \ answerable \\
NA, & if \ q_i \ is \ unanswerable \\
\end{array}
\right.$$

\section{Methodology}
\label{sec:method}

We now introduce how we combine discourse structure to represent multiparty dialogue with a neural network. The architecture of our model is shown in Figure~2. Our model consists of three parts: sequential context encoding, discourse graph modeling, and MRC module. The sequential context encoding module aims to learn the sequence structure of utterances. The discourse graph modeling module constructs the multiparty discourse graph using discourse dependency links and discourse relations. Finally, the MRC module finds the answer span, where applicable.





\subsection{Pre-processing: utterance encoding}

Different from the traditional MRC task, the input of a multiparty dialogue consists of a sequence of utterances originating from different speakers. 
We first encode the representations of utterances as the input of our model.

In the related work of DialogueGCN, their model adopts the Convolution Neural Network \cite{kim2014convolutional} to learn the representation of each utterance, 
using 
a single convolutional layer followed by max-pooling and a fully connected layer. 

In contrast to DialogueGCN's modeling decision, we adopt the widely-used pretrained model BERT to extract features $u_i$ of utterances. 
Fig.~3 shows the BERT input representations. 
We adopt the \texttt{[CLS]} from well-trained BERT model as the representations of utterances as the inputs of our model.  
To be clear, 
the utterance encoder does not participate in the model training; 
BERT's CLS model is employed to obtain an encoded representation of each utterance.
    
\begin{figure}
    \centering
    \includegraphics[width=0.45\textwidth]{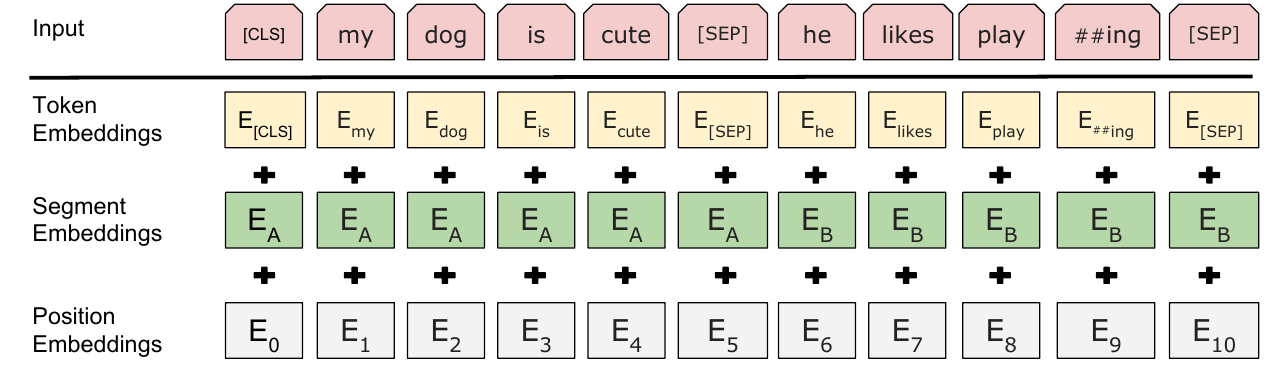}
    \caption{ BERT input representations \cite{devlin2019bert}. The input embeddings are the sum of the token embeddings, the segmentation embeddings and the position embeddings.}
    \label{fig:bert}
\end{figure}

\subsection{Sequential context encoding}

The sequential context encoder models the dialogue structure according to the timeline of utterances, regarding dialogue as a sequence of utterances. 
This module learns the sequential structure of utterances in input dialogue and outputs the new representations of utterances.
Inspired by DialogueGCN~\cite{ghosal2019dialoguegcn}, after obtaining the context-independent representation $u_i$ of each utterance, we model the sequential structure of dialogues by the Bi-directional GRU (Bi-GRU) using Equation~(1) to learn the context-dependent representation of each utterance. 
\begin{equation}
g_i=BiGRU(g_{i(+,-)1},u_i)
\end{equation}

Our choice of a bidirectional GRU model is modular; it can be easily replaced by other sequential modeling encoders, such as other recurrent neural network architectures or a transformer model.

In a multiparty dialogue, discourse relations can exist between two distant utterances and are substantially affected by long-distance dependencies. 
Therefore we must augment the discourse relation detection from just adjacent utterances (sequence), and also apply it to non-adjacent utterances (a graph).  We construct a dialogue discourse graph in the next module. 


\subsection{Discourse graph modeling}

This module is a graph neural network that aims to learn the dialogue discourse graph using Graph Convolutional Network (GCN)\cite{defferrard2016gcn}, addressing the modeling of discourse dependency links and discourse relation types in conversation.

\subsubsection*{Graph construction}

The outputs of sequential context encoder are context-aware utterances representations 
$\{g_1,g_2,...,g_i,...g_N\}$ that are inputs of dialogue discourse graph modeling module. For graph construction, 
each utterance $u_i$ is regarded as a vertex in the directed graph $G=(V, E, R)$ where $V$ is the vertex set, $E$ is the edge set, $R$ is the relation set. 

\paragraph{Vertices.} In the dialogue discourse graph, each utterance $u_i$ is represented as a vertex $v_i$. In Figure~2, five vertices represent five utterances from three different speakers, shown in different colors. We assume that all vertices in a dialogue graph are connected (i.e., one large graph component; no isolated nodes).

\paragraph{Edges.} We adopt discourse dependency links as the directional edges in the dialogue discourse graph. An edge means that there is a discourse dependency relation between the two utterances. For instance, if utterance $u_j$ depends on utterance $u_i$, there would be an edge of $e_{ij}$. As the discourse graph is directional, $e_{ij}$ is not equivalent to $e_{ji}$. In the majority of cases, an utterance only depends on its previous utterances, so the direction of edges are 
often directed as a topological sort from earlier utterances to later ones. In training, since all edges are from the ground truth in Molweni, we do not distribute weights for each edge.

In the DialogueGCN, as there are no discourse information in the dataset, the speaker-level context encoder models an utterance using its previous ten and the following ten utterances to construct a fully connected graph within a window context. 
Different from DialogueGCN that constructs a fully connected graph within a context utterance window, our model introduces the dialogue's discourse structure: 
directional discourse links represent the discourse dependency link, which is also associated with a specific relation type.
The golden annotation of discourse structure is provided during training and testing.

As seen in Fig.~2, there are only five edges among the five vertices. According to the statistics of the STAC corpus \cite{D15DP4MultiPparty}, each utterance participates in 1.06 discourse relations with other utterances, on average. Therefore, the discourse dependency graph is very sparse; it is mostly a chain. Constructing an appropriate dialogue graph using discourse structure can reduce computing costs, compared to using the sliding window, fully connected graph. The training time and GPU memory use for DialogueGCN are two and four times greater respectively, compared to our model, as empirically measured in our experiments. 
In Fig.~2, we use a solid line to denote the discourse dependency between utterances from different speakers and use a dotted line to represent dependencies between utterances from one speaker. 

\paragraph{Relations.} The relations on the edges are discourse relation types. For example, $r_{ij}$ is the discourse relation type edge $e_{ij}$ which is the discourse dependency link between utterance $u_j$ and utterance $u_i$. We adopt the discourse relation hierarchy from STAC \cite{asher2016discourse}, which includes 16 types of discourse relations: \textit{Comment}, \textit{Clarification\_question}, \textit{Elaboration}, \textit{Acknowledgement}, \textit{Continuation}, \textit{Explanation}, \textit{Conditional}, \textit{Question-Answer\_pair (QAP)}, \textit{Alternation}, \textit{Question-Elab(Q-Elab)}, \textit{Result}, \textit{Background}, \textit{Narration}, \textit{Correction}, \textit{Parallel} and \textit{Contrast}. In Fig.~2, the color of edges represents the discourse relation types. In the example, there are three different discourse relations: \textit{Elaboration}, \textit{Question--Answer\_pair}, and \textit{Acknowledgement}.

\subsubsection*{Graph representation}

To construct the graph structure of the dialogue, DADgraph models each utterance according to a given discourse structure. We use $g_i$ to initialize $v_i$ which is obtained from the sequential context encoder and includes utterance features.

We introduce $H_i$ to compute features of utterance $u_i$ by aggregating utterances which have discourse dependency relations.
\begin{equation}
\begin{aligned}
h_i^{(1)}&=\sigma(\sum_{r\in R}\sum_{j \in N_i^r}  \frac{\alpha_{ij}}{c_{i,j}}W_r^{(1)} +\alpha_{ii}W_{0}^{(1)}g_i) \\
h_i^{(2)}&=\sigma(\sum_{r\in R}W^{(2)}h_j^{(1)}+W_0^{2}h_i^{(1)})
\end{aligned}
\end{equation}
\noindent where $h_i^{(1)}$ and $h_i^{(2)}$ are new feature vectors computed by aggregating other utterances with their discourse dependency links, using the output of sequential encoder $g_i$ as inputs.


After running the discourse graph modeling module, we obtain a new, augmented feature representation of vertex $v_i$ (from $g_i$) $h_i$, which incorporates information about its neighborhood in the directional discourse graph.

\begin{figure}
    \centering
    \includegraphics[width=0.45\textwidth]{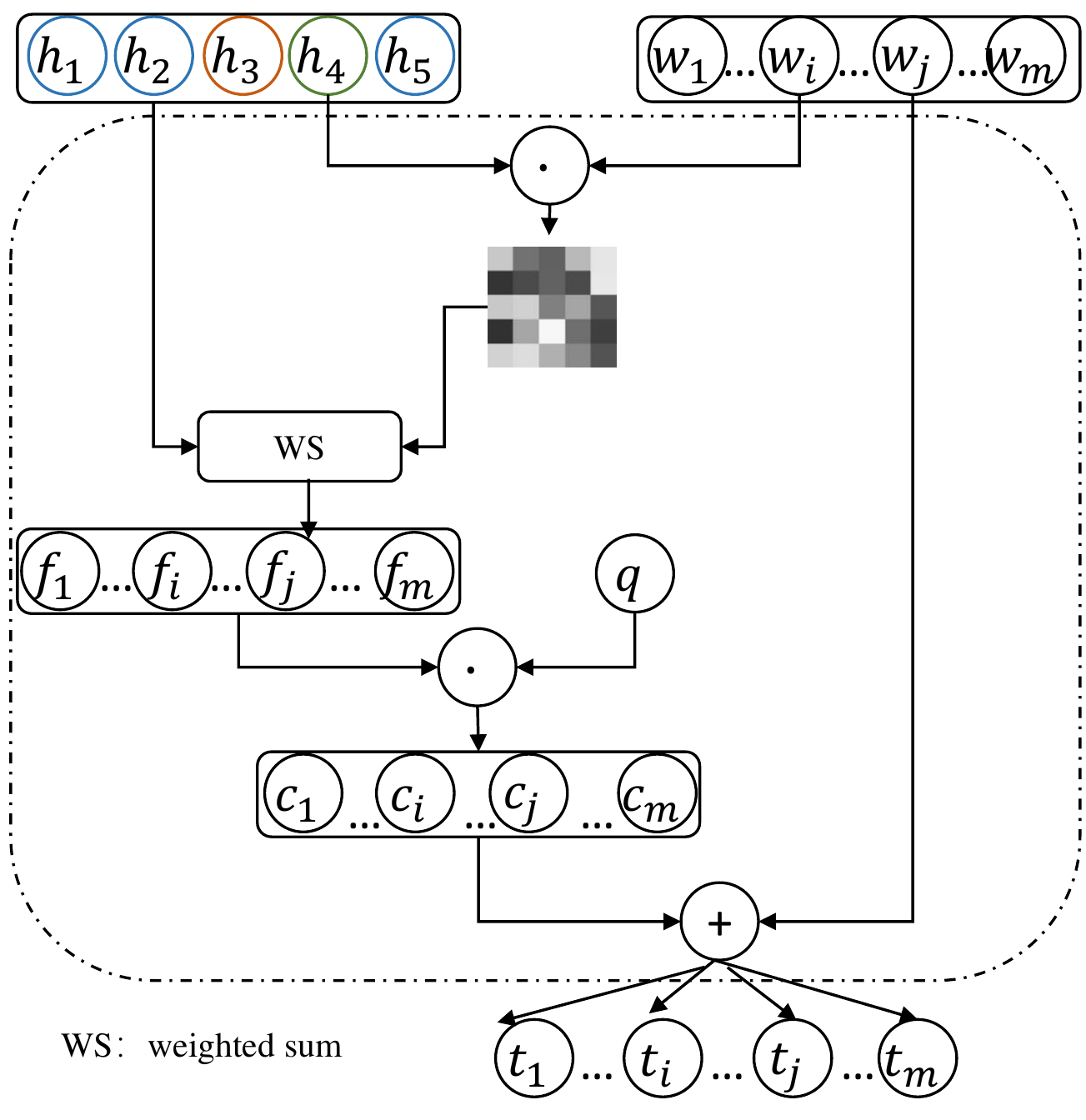}
    \caption{The interactions in our MRC module. $w_i$ are the representations of word $i$ in the dialogue, and $q$ is the question. WS denotes weighted sum.}
    \label{fig:mrc}
\end{figure}

\subsection{MRC Module}

In the MRC module, we receive utterance representations $h_i$ from the discourse graph modeling module, as well as the embeddings of each word $w_i$ in source dialogue, and the representation of question $q$ as inputs. MRC then outputs the answer span $(S, E)$ of question $q$ in the dialogue, when the question is inferred as answerable; $NA$ if inferred as unanswerable.
The interactions among words, utterances, and the question are shown in Fig.~4.

MRC combines the utterance representations to word representation via attention to introduce the dialogue discourse graph structure to all words. Based on the discourse-aware word representations, our MRC module predicts whether a word can be the start or end of an answer. We adopt simple interaction between dialogue and the question, so we can analyze the effect of dialogue graph modeling with discourse structure; future work could examine more sophisticated interactions. 

We first compute the interaction between words $w_j$ of input dialogue and utterance representation $h_i$ obtained from speaker-level context encoder and obtain the attention weighted $\alpha_{ij}$. We then compute the weighted sum for aggregating attention scores as the weight of each utterance and obtain new features of each word $f_i$. In this case, $f_i$ is regarded as the combination of word feature and discourse structure of utterances.
\begin{equation}
\begin{aligned}
e_{i,j}&=h_i \cdot w_j \\
\alpha_{ij}&=\frac{exp(e_{ij})}{\sum_{k=1}^{M}exp(e_{ik})}\\
f_i&=\sum_{j=1}^{N}\alpha_{ij}h_i\\
\end{aligned}    
\end{equation}

To answer the given question $q$, we perform the dot product between $f_i$ and $q$. The obtained new representation $c_i$ thus considers the question information for each word in the dialogue.
\begin{equation}
\begin{aligned}
c_i&=f_i \cdot q\\        
\end{aligned}
\end{equation}

Finally, we concatenate the source word embeddings and weighted utterance embedding as final word embeddings. Therefore, for each word of input dialogue $d$, $t_i$ contains two sources of information: word features and question-aware representations with discourse dependencies of utterances.
\begin{equation}
    \begin{aligned}
    t_i&= concat(w_i, c_i)\\        
    \end{aligned}
\end{equation}

To find the answer span in the input dialogue, we introduce a start vector $S \in \mathbb{R}^H$ and an end vector $E \in \mathbb{R}^H$.  These respectively represent the start and end of the answer span. We compute the probability of each word $i$ being the answer span and candidate span is computed as follows:
\begin{equation}
\begin{aligned}
s_{NA}&=S \cdot C + E \cdot C    \\
s_{i,j}&=max_{j\geq i}S\cdot t_i+E\cdot t_j    
\end{aligned}
\end{equation}
\noindent where $C$ is the vector of representing all words in the dialogue. $s_{NA}$ and $s_{i,j}$ is respectively the probability of the question $q$ being unanswerable and best non-null answer span $(i,j)$. If $s_{i,j}>s_{NA}+\tau$, we predict the question is answerable and the span $(i,j)$ is the answer. Hyperparameter $\tau$ thus controls the system's necessary confidence level to declare a specific answer.

\section{Experiments}
\label{sec:exp}

\subsection{Dataset: Molweni}


\begin{table}
    \begin{center}
        \begin{tabular}{l|l|c|c|c}
            \hline \bf   \bf   & \bf Train  & \bf Dev & \bf Test & \bf Total \\ 
            \hline
            Dialogues      & 8,771   &  883 & 100 & 9,754 \\
            Utterances     & 77,374   &  7,823 &  845 & 86,042\\ 
            Questions     & 24,682 & 2,513 & 2,871 & 30,066  \\
            \hline
        \end{tabular}
    \end{center}
    \caption{\label{font-table} Overview of Molweni for MRC.}
    \vspace{-0.16in}
\end{table}

With the exception of Molweni, no multiparty dialogue dataset for MRC annotates the discourse structure of dialogues.  Also currently, the state-of-the-art performance of off-the-shelf dialogue discourse parsers is still unsatisfactory. 
In this paper, we perform experiments on the \textit{Molweni} dataset. The overview of the Molweni dataset is shown in Table~1. 

Considering the properties of multiparty dialogues, the Molweni dataset is presented, a machine reading comprehension (MRC) dataset built over multiparty dialogues. 
Molweni dataset derives from the large-scale multiparty dialogues dataset the Ubuntu Chat Corpus \cite{lowe2015ubuntu}, which is a large-scale multiparty dialogues corpus. 
To learn better graph representations of multiparty dialogues, Molweni adopts the dialogues with 8--15 utterances and 2--9 speakers. To simplify the task, the dataset filters out the dialogues containing long sentences (more than 20 words). Finally, Molweni randomly chooses 10,000 dialogues with 88,303 utterances from those that qualify from the Ubuntu dataset.

\subsection{Baselines and evaluation}

We use two kinds of models as experiment baselines: classic MRC models for passages, and models that representing multiparty dialogues. 

\paragraph{MRC models for passages understanding} We adopt three well-known MRC models that can answer unanswerable questions as baselines:
\begin{itemize}
    \item \textbf{BiDAF \cite{seo2016BiDAF}}. The BiDAF model presents the context passage at different levels of granularity and learns the query-aware context representation using a bi-directional attention flow mechanism.
    \item \textbf{DocQA \cite{clark2018simple}}. This model is a neural paragraph-level QA method, which can scale to document and multi-document inputs. 
    DocQA can ignore no-answer containing paragraphs in documents. 
    The model contains paragraph sampling and attempts to produce a globally correct answer.
    \item \textbf{BERT \cite{devlin2019bert}}. BERT is a bidirectional encoder utilizing   transformers \cite{devlin2019bert}. To learn better representations for text, BERT adopts two objectives: masked language modeling and the next sentence prediction during pretraining. To adapt BERT for our task, we concatenate all utterances from the input dialogue as a passage, 
    where each utterance $u_i$ encodes both the speaker identity and their uttered text as $\{speaker_{u_i}: content_{u_i}\}$.
\end{itemize}

\paragraph{Neural Models for Dialogue Modeling} We adopt DialogueRNN \cite{majumder2019dialoguernn}and DialogueGCN \cite{ghosal2019dialoguegcn}as our baselines.  These two models are originally designed for sentiment classification.  To adapt them to our task, we replace DADgraph's internal models with these models, but hold fixed the same final MRC module and BERT-based utterance representations.
\begin{itemize}
    \item \textbf{DialogueRNN}. DialogueRNN is a sequential neural network model for representing multiparty dialogues on emotion recognition for conversations task with two bi-directional GRUs: a global GRU and a party GRU.
    \item \textbf{DialogueGCN}. Compared to DialogueRNN, DialogueGCN model the context windows of an utterance in the dialogue as a graph and represent the graph using the GCN model.
\end{itemize}

\paragraph{Evaluation metric and upper bounds} Our task is closely related to SQuAd~2.0, so we adopt the same evaluation metrics: exact match (EM) and \textit{$F_1$} score to evaluate experiments. EM measures the percentage of predictions that match all words of the ground truth answers exactly. $F_1$ scores are usually engineered to be more tolerant, measuring the average overlap between a system's prediction and a ground truth answer. We ask two volunteers that have a computer science background and who understand technical dialogues well to answer questions in the test set.  Our interannotator study indicates that our volunteers achieved 64.3\% in EM and 80.2\% in $F_1$ score on the Molweni dataset. 

\subsection{Results}

\begin{table}
    \begin{center}
        \begin{tabular}{l|l|c}
            \hline \bf   \bf   & \bf EM  & \bf F1 \\
            \hline
BiDAF \cite{seo2016BiDAF} & 22.9 & 39.8\\
DocQA \cite{clark2018simple}& 42.5 & 56.0\\
BERT \cite{devlin2019bert}  & 45.3 & 58.0\\
DialogueRNN \cite{majumder2019dialoguernn} & 45.4 & 60.9\\
DialogueGCN \cite{ghosal2019dialoguegcn}  & 45.7 & 61.0\\
\hline
DADgraph (Our)  & \textbf{46.5} & \textbf{61.5}\\
\hline
Human performance  & 64.3 & 80.2 \\
\hline
\end{tabular}
\end{center}
\caption{\label{font-table} Results on Molweni dataset.}
\end{table}

Table~2 shows the results on Molweni. BiDAF achieves the lowest results in both EM and $F_1$ measures, and the DocQA model obtains improvements compared to the BiDAF model. As expected, both models do not perform well compared against other models, because two models are designed to model passages understanding which is quite different from multiparty dialogue understanding.  BERT is a strong baseline for representing passages, bettering both BiDAF and DocQA. We observe that DialogueRNN and DialogueGCN achieve higher results compared to the BERT model on both EM and \textit{$F_1$} measures.  This signifies that such dialogue-based models can learn better representations for dialogues than BERT, and that such represention is important to MRC performance. We also note a genre discrepancy: BERT is pretrained on well-written passages, quite different from dialogue text.

Our DADgraph, which employs ground truth discourse structure achieves the best results. First, compared to BiDAF, DocQA, and BERT, our dialogue-based model yields improved results that showcase the efficiency of the dialogue-based representation learning model. Second, compared to other dialogue-based models, our model demonstrates that discourse-awareness can create improved representations that better reflect the semantic relations among utterances. As a side effect, DADgraph's model incurs less memory and time costs compared against DialogueGCN, as DialogueGCN adopts a sliding window method and  constructs a fully connected graph.




\subsection{Ablation Study}

We perform ablation experiments to verify the effect of discourse dependency links and discourse relation types. The results of ablation experiments on the Molweni dataset are shown in Table~3. 

\paragraph{Evaluation discourse relation types.} To verify the influence of discourse relation types, we replace discourse relation with relations in 
vanilla dialogue which depends on two aspects: speaker dependency and temporal dependency. 
For example, when utterance $u_i$ and $u_j$ co-occur in a conversation, this ablated model does not consider whether $u_i$ is uttered before $u_j$ or after (a bag-of-utterance assumption). From Table~5, when removing discourse relation types, both EM and $F_1$ results decrease. 

Our ablation experiments
indicate the effect of discourse relations on understanding dialogues. Discourse relations are helpful to understand the dialogue and find the correct span from the dialogue.

\begin{table}
    \begin{center}
        \begin{tabular}{l|l|c}
            \hline \bf   \bf   & \bf EM  & \bf F1 \\ 
            \hline
            DADgraph      & 46.5   &  61.5 \\
            - w/o discourse relations     & 44.9   &  60.6 \\ 
            - w/o discourse structure      & 44.7   &  60.5 \\ 
            \hline
        \end{tabular}
    \end{center}
    \caption{\label{font-table} Results of ablation experiments on Molweni dataset.}
    \vspace{-0.16in}
\end{table}

\paragraph{Evaluation on discourse structure.}
To verify the help of discourse structure, we adopt a fully connected structure to build an utterance dialogue graph.
When using a fully-connected utterance window graph, no corresponding discourse relations to edges are provided in our dataset. Therefore, we only can evaluate the influence of discourse structure including both links and relations. 
From Table~5, when removing both discourse links and relation and adopt a fully connected graph to represent the dialogue, EM and $F_1$ results all decrease. 
Ablation experiment results prove the help of discourse structure for modelling dialogues.

\begin{figure*}
    \centering
    \includegraphics[width=\textwidth]{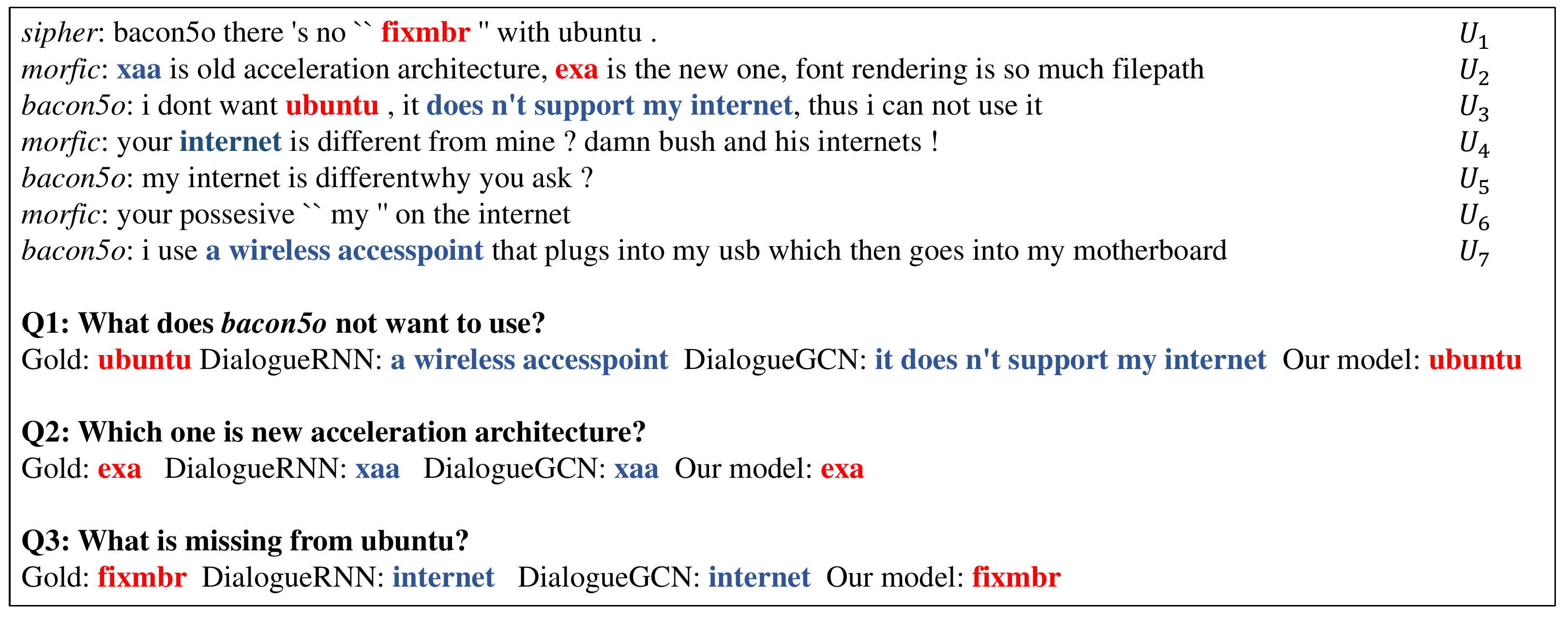}
    \caption{ An example from our Molweni test set with three speakers: \textit{sipher}, \textit{morfic}, and \textit{bacon5o}. Q1,Q2 and Q3 are three answerable questions in test set. We show ground truth answers and the output answers of DialogueRNN, DialogueGCN, and our model. Correct answers are marked in red; incorrect ones in blue.}
    \label{fig:case}
    \vspace{-0.20in}
\end{figure*}

\subsection{Case study}

In this part, we analyze a dialogue from Molweni where DADgraph correctly answers the questions given the discourse structure that DialogueRNN and DialogueGCN baselines yields incorrect answers. Figure~4 shows an example from the Molweni test dialogues with two answerable questions. In the dialogue, there are three speakers and seven utterances. 

The first question that we examine is ``What does \textit{bacon5o} not want to use?". The answers of DialogueRNN and DialogueGCN for Q1 are ``a wireless accesspoint" and ``it does n't support my internet", respectively. 
DialogueRNN only models the sequential structure of utterances using the RNN method, which would be limited to long-term dependency problems. 
Different from DialogueRNN, DialogueGCN can construct a dialogue graph that can be used to model semantic relations between long-distance utterances, but the way of constructing a fully connected graph does not accurately obtain the structure information in the dialogue and pay huge computing costs. 



The second question (Q2) is ``which one is the new acceleration architecture?". In the dialogue, there are two acceleration architectures mentioned: $xaa$ and $exa$. Considering the occurrence of ``acceleration architecture", both DialogueRNN and DialogueGCN output the incorrect answer $exa$ for Q2. 

The third question (Q3) is ``What is missing from ubuntu?". The word ``ubuntu'' in the question is an important clue for finding the answer. The word ``ubuntu" appears in both $U_1$ and $U_3$. However, both DialogueRNN and DialogueGCN output the answer $internet$ for question Q3, which is incorrect, originating from $U_3$. 

In Fig. 5, our DADgraph correctly answers these three answerable questions. The complex structure of multiparty dialogues makes it difficult to understand them.  
After introducing discourse structure, our model can learn better representations of each utterance and adopt the structure to find the index of start and end of answers. 

\section{Conclusion}

In this paper, we propose a discourse-aware dialogue graph neural network, \textit{DADgraph}, for multiparty machine reading comprehension tasks.  It features a pipeline of three components: sequential context encoding, dialogue discourse graph modeling, and an MRC module. To the best of our knowledge, our model first introduces the discourse structure on multiparty dialogues MRC tasks. To verify the performance of our model, we perform experiments on the Molweni corpus, a large-scale multiparty dialogues dataset for MRC with discourse structure.

Our experimental results on the Molweni dataset show that discourse structure helps understand the dialogue compared with traditional MRC models on passage and pretrained language models. 
\section*{Acknowledgements}

We thank anonymous reviewers for their helpful comments. Thanks to Yibo Sun and Tianwen Jiang for their advice for this paper. The research in this article is supported by the Science and Technology Innovation 2030 - "New Generation Artificial Intelligence" Major Project (2018AA0101901), the National Key Research and Development Project (2018YFB1005103), the National Science Foundation of China (61772156, 61976073), Shenzhen Foundational (JCYJ20200109113441941), Research Funding and the Foundation of Heilongjiang Province (F2018013).




\bibliography{mrc}
\bibliographystyle{IEEEtran}

\end{document}